\documentclass[a4paper,conference]{IEEEtran}

\usepackage{graphicx}
\usepackage{amssymb}
\usepackage{mathtools}
\usepackage{array}
\usepackage[misc,geometry]{ifsym}
\usepackage{color}
\usepackage{breqn}
\usepackage{xspace}

\usepackage{stfloats}
\newcommand\red[1]{\color{black}#1\xspace\color{black}}
\newcommand\green[1]{\color{black}#1\xspace\color{black}}

\usepackage{color, hyperref}
\hypersetup{
    colorlinks=true,
}
\begin{document}

\title{Font Generation with Missing Impression Labels}

\author{\IEEEauthorblockN{Seiya Matsuda}
\IEEEauthorblockA{Kyushu University\\
Fukuoka, Japan\\
seiya.matsuda@human.ait.kyushu-u.ac.jp}
\and
\IEEEauthorblockN{Akisato Kimura}
\IEEEauthorblockA{NTT Communication Science Laboratories\\
NTT Corporation, Kanagawa, Japan\\
akisato@ieee.org
}
\and
\IEEEauthorblockN{Seiichi Uchida}
\IEEEauthorblockA{Kyushu University\\
Fukuoka, Japan\\
uchida@ait.kyushu-u.ac.jp
}}


%


\maketitle

\begin{abstract}
Our goal is to generate fonts with specific impressions, by training a generative adversarial network with a font dataset with impression labels. The main difficulty is that font impression is ambiguous and the absence of an impression label does not always mean that the font does not have the impression.
This paper proposes a font generation model that is robust against missing impression labels. The key ideas of the proposed method are (1)~a co-occurrence-based missing label estimator and (2)~an impression label space compressor. The first is to interpolate missing impression labels based on the co-occurrence of labels in the dataset and use them for training the model as completed label conditions. The second is an encoder-decoder module to compress the high-dimensional impression space into low-dimensional. We proved that the proposed model generates high-quality font images using multi-label data with missing labels through qualitative and quantitative evaluations. Our code is available at
\href{url}{https://github.com/SeiyaMatsuda/Font-Generation-with-Missing-Impression-Labels}.

\end{abstract}

%
\IEEEpeerreviewmaketitle

\section{Introduction\label{sec:intro}}
\begin{figure}[t]
    \centering
    \includegraphics[width=1.0\linewidth]{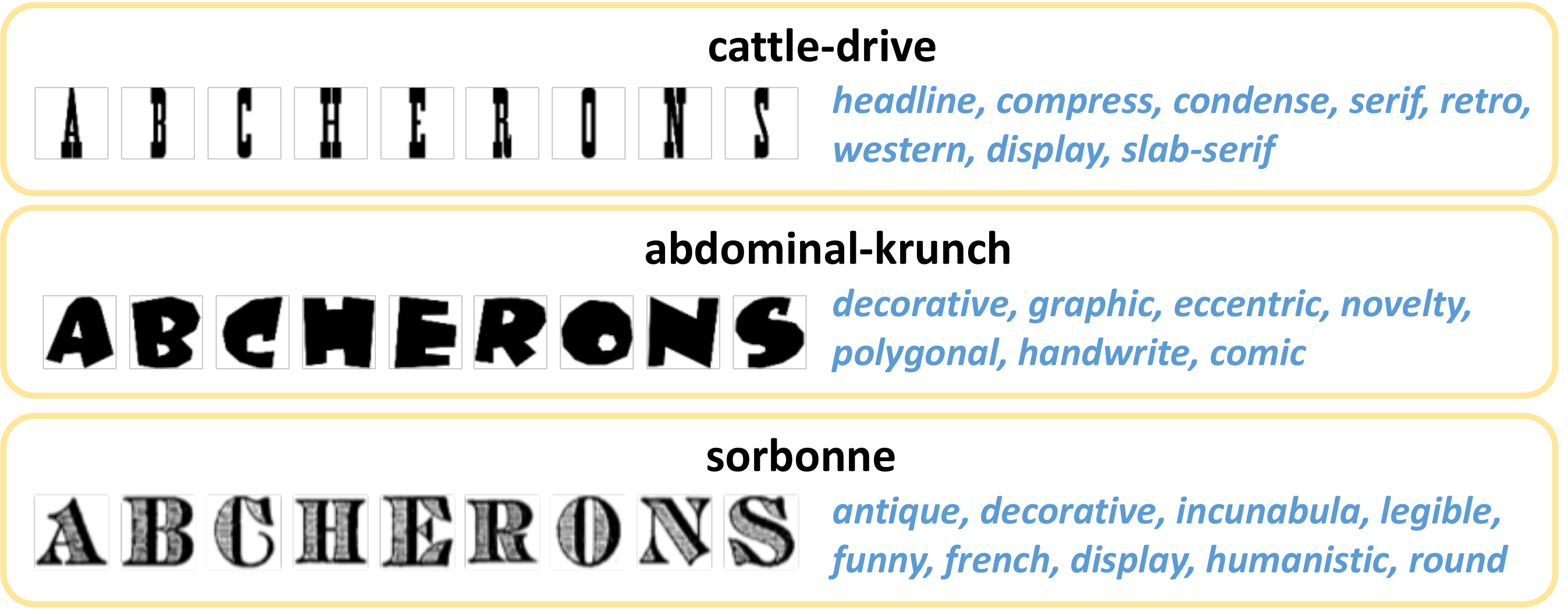}\\[-3mm]
    \caption{Three font examples and their impression labels from MyFonts dataset\cite{Chen2019large}.
    The font {\tt abdominal-krunch} could have the impression labels {\it thick} and/or {\it bold}; however, none has attach them.}
    \label{fig:myfonts}
\end{figure}

Impressions of fonts enrich typographic designs, but they are subjective and often ambiguous.
Fig.~\ref{fig:myfonts} shows three fonts and their impression labels from the MyFonts dataset~\cite{Chen2019large}. The impression labels are attached by crowdsourcing; various font experts and non-experts freely attach the labels to each font.
\red{Moreover, the impression labels are open-vocabulary; there is no pre-defined list of impression labels.}
Consequently, impression labels \green{attached} are often incomplete. For example, {\tt abdominal-krunch} in Fig.~\ref{fig:myfonts} could have the impression labels {\it thick} and  {\it bold}.  On the other hand, it is too optimistic to expect the complete impression labels, by considering the ambiguity in impressions. In other words, it is difficult to determine unanimously whether a certain font has a certain impression. \par
When using this font-impression dataset~\cite{Chen2019large} for various applications, the ambiguity of the impression labels, especially {\em missing labels}, will be a severe problem. \red{More specifically, for two very similar fonts, an impression label may be attached to only one of them, but not the other.}  In a multi-label classification task, \green{such missing labels} should {\em not} be treated as 
a negative label during training a classifier. In the above example, the classifier should not be trained by considering {\tt abdominal-krunch} as a negative sample of the {\it thick} class. 
In recent years, the problem of missing labels has been actively researched in multi-label classification~\cite{yu2014large,  song2020learning, cole2021multi}.\par
The purpose of this paper is to generate font images with a specific impression by a conditional  generative adversarial network (GAN), trained by the font-impression dataset with missing labels. 
When we train the conditional GAN, impression labels are used as the condition. 
Missing labels result in incomplete conditions, and therefore some mechanisms to complete missing labels are necessary. To the authors' best knowledge, this is the first attempt to deal with the missing label problem in a conditional image generation task.\par
\begin{figure*}[t]
    \centering
    \includegraphics[width=\linewidth]{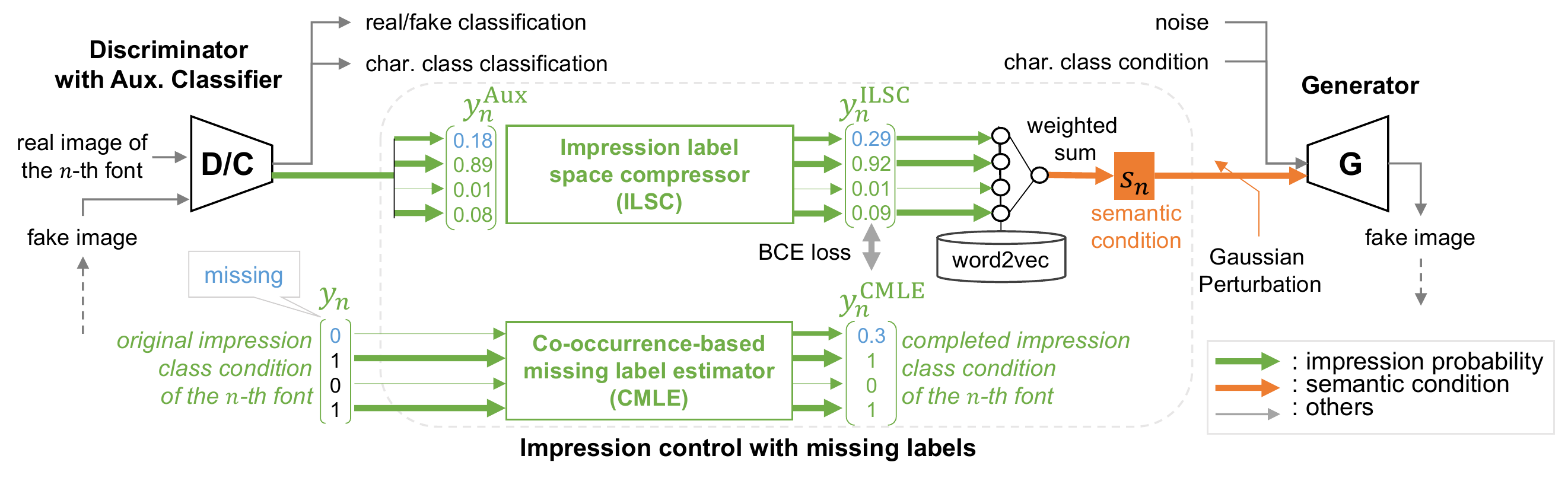}\\[-3mm]
   \caption{The overall structure of the proposed model. The two modules, impression label space compressor (ILSC) and co-occurrence-based missing label estimator (CMLE), are highlighted because they are newly introduced for missing labels. }
    \label{fig:proposed_model}
\end{figure*}

Fig.~\ref{fig:proposed_model} shows the overall architecture of the proposed model. The proposed model uses Classifier's Posterior GAN (CP-GAN)~\cite{kaneko2019CP-GAN}, a robust conditional GAN model, as a baseline and has two novel components to cope with missing labels. The first component is a {\em co-occurrence-based missing label estimator} (CMLE), which provides a completed impression class condition $y^{\rm CMLE}_n$ from a given (incomplete) original impression class condition $y_n$ by statistical estimation. The second component is an {\em impression label space compressor} (ILSC), which embeds a given label condition into a learned low-dimensional space, where the probabilities of missing labels are boosted. Our code is available at
\href{url}{https://github.com/SeiyaMatsuda/Font-Generation-with-Missing-Impression-Labels}.
\par

The main contributions of this paper can be summarized as follows. 
\begin{itemize}
    \item This paper proposes a novel conditional GAN model for generating font images with a specific impression. The technical highlight is its robustness to missing labels, which is an inevitable problem for dealing with ambiguous impressions.  
    \item To the authors' best knowledge, this is the first attempt at realizing an image generation system with missing labels. Although we focus on font image generation with impression class conditions, the idea of the proposed method is applicable to different image generation tasks with missing labels. 
    \item Quantitative and qualitative experiments demonstrate that the proposed method can generate high-quality font images with various valuations, compared to the other methods. This result supports the robustness of the proposed method to missing labels.
\end{itemize}

\section{Related Work}
\subsection{GAN based font generation}\par
Recently, font generation has been greatly advanced by developing Generative Adversarial Networks (GANs)~\cite{goodfellow2014generative}. Azadi et al.~\cite{azadi2018multi} proposed a font interpolation method to generate a set of multi-content fonts from a small number of character classes according to a consistent style. Style transfer-based methods are also very widely used. zi2zi~\cite{Kaonashi2017} proposed synthesizing Chinese calligraphy images, while controlling them using category information. Xie et al.~\cite{xie2021dg} proposed a new deformable DGFont for unsupervised font generation. Vector glyph generation has also been researched. 
Hayashi et al.~\cite{hayashi2019glyphgan} proposed GlyphGAN, which is based on DCGAN~\cite{radford2015unsupervised} and can control the font style while maintaining character style consistency.
Wang et al.~\cite{wang2021deepvecfont} proposed DeepVecFont, which generates font images in a vector format.\par
Our goal is to generate fonts under a specific impression condition. 
However, such a conditional font generation method is still not common. Wang et al.~\cite{wang2020attribute2font} proposed Attribute2Font, which synthesizes a font from specific attributes ($\sim$ impressions). This paper uses the dataset created by O'Donovan et al.~\cite{o2014exploratory} to synthesize an appropriate font. However, the number of attributes is limited to 37. Matsuda et al.~\cite{matsuda2021impressions2font} proposed Impressions2Font(Imp2Font, for short), which generates fonts directly from impression labels. 
Since a word-embedding mechanism is used to convert impression labels into a real-valued condition vector, Imp2Font realizes robustness to the noisy impression labels. However, the problem of missing impression labels is not taken into account. In our trial, we propose a font generation method trained while completing the missing labels as much as possible. The proposed method is the first trial that takes into account the problem of the missing 
label in the conditional generation model.

\subsection{Font style and impression}
The relationship between font and its impression has been analyzed from the 1920's~\cite{poffenberger1923study} via subjective and psychological experiments. Nowadays, we can find several attempts to analyze the relationship as computer science research.  O’Donovan et al.~\cite{o2014exploratory} used crowdsourcing to collect font attribute data. The attributes include impressions of font shapes such as {\it legible} and {\it warm}, and real-values are given for each attribute. However, the number of attributes is limited to 37. Wang et al.~\cite{wang2015deepfont} applied a convolutional neural network (CNN) to the font classification task. Shirani et al.~\cite{Shirani2020} analyzed the relationship between the visual attributes of a font and the linguistic context of the text to which the font is usually applied.\par
Chen et al.~\cite{Chen2019large} have released a dataset collected from MyFonts.com. In this dataset, multiple impression labels (e.g., {\it elegant}, {\it pretty}) are given to each of the 18,815 fonts. As noted in Section~\ref{sec:intro}, the impression labels were attached by crowdsourcing and thus often incomplete. However, this database allows us to conduct a large-scale analysis on the font shape-impression relationship. For example, Ueda et al.~\cite{ueda2021parts} used this dataset to analyze the correlation between local shapes and impressions. Kang et al.~\cite{kang2021shared} examined the cross-modal embedding of font images and their impressions into a shared latent space. 

\subsection{Leaning with missing labels}
Multi-label classification is a well-studied problem in machine learning research. In multi-label classification tasks, we often encounter datasets with missing labels. For example, Yu et al.~\cite{yu2014large} employ label compression-based methods to compensate for missing labels. The problem of missing labels is also similar to learning from positive and unlabeled data (i.e., so-called PU-learning). For this situation, Kanehira et al.~\cite{kanehira2016multi} extended ranking learning to multi-label classification to reduce the negative impact of label incompleteness. Also, Cole et al.~\cite{cole2021multi} considered a setting where only one positive label is given and succeeded in approaching the performance of a complete labeled case.\par
However, there is little work that considers missing labels in generative models. Guo et al.~\cite{guo2020positive} proposed a method to stabilize the learning of the discriminator by considering the positive, unlabeled classification problem of GAN. However, the purpose of their method is to stabilize learning and thus not to complete missing labels in a conditional generation. 
\section{MyFonts Dataset\label{sec:dataset}}
We use the MyFonts dataset~\cite{Chen2019large} in all our experiments. Fig.~\ref{fig:myfonts} shows three font examples with their impression labels.  We use 26 char classes (26 English capitals, `A'-`Z'). From the dataset, we removed dingbats (i.e., unreadable illustrations) by manual inspections. We removed impression labels that are not in the vocabularies of word2vec, which was pre-trained by the Google News Dataset. As a result, we use $N=17,202$ fonts (among 18,815) and $K=1,430$ impression labels (among 1,824) in the experiment. The maximum, minimum, and average numbers of impression labels for each font are 184, 1, and 14.39, respectively. \par

\section{The proposed model}
\subsection{Overall structure}
Fig.~\ref{fig:proposed_model} shows the proposed model for generating fonts with specific impressions. As presented in Section~\ref{sec:intro}, the proposed model follows the framework of Classifier's Posterior GAN (CP-GAN)~\cite{kaneko2019CP-GAN}, a multi-label conditional GAN model with auxiliary classifiers, as a baseline. As with standard conditional GANs, its generator receives the character class condition represented by a 26-dim one-hot vector and the original impression class condition represented by a $K$-dim multi-hot vector ($y_n$ in Fig.~\ref{fig:proposed_model}). Note that the former is believed to be complete, but  the latter might contain positive but missing labels. The generator of CP-GAN employs the soft prediction by the auxiliary classifier as a conditional input, which enables us to capture between-class relationships and generate an image selectively conditioned on the class specificity. Our model additionally introduces a word embedding layer to derive a semantic condition $s_n$ that is fed to the generator. This semantic condition enables us to generate images considering  the semantic similarity of labels. \par

CP-GAN asks the auxiliary classifier to emit the output as close to the original impression class condition $y_n$ as possible. However, this might damage the quality of generated images, since our problem setting indicates that the original impression class condition $y_n$ might contain missing labels. Eventually, the model will suppress impressions related to missing labels for generating font images. \par
Our novel components, co-occurrence-based missing label estimator (CMLE) and impression label space compressor (ILSC), solve the above issue. The former is a static procedure to transform an original impression class condition $y_n$ to 
a completed one $y^{\rm CMLE}_n$ in advance to train 
the GAN.  The latter is dynamically trainable with the generator and discriminator to make the internal condition $y^{\rm Aux}_n$ from the auxiliary classifier ``smoother'' to boost the probability of missing labels. We will describe them in the following sections in detail.

\subsection{Co-occurrence-based missing label estimator}\par
CMLE transforms the original impression class condition  $y_n$
into a completed impression class condition $y^{\rm CMLE}_n$. 
This transformation is determined as a pre-defined matrix $T$, which represents the co-occurrence between two impression labels.  Specifically, $T$ is defined and fixed as follows:
The binary variable $y_{n,i}\in\{0,1\}$ shows whether the $i$-th impression label is attached (1) or not attached (0) to the $n$-th font image, where $n\in \{1,\ldots, N\}, i\in \{1,\ldots, K\}$. Then, the matrix $T$ is defined as $T=(T_{ij})\in[0,1]^{K\times K}$, where $T_{ij}=P(j|i)$ is the conditional probability and given as follows:
 \begin{equation}
 T_{ij}=P(j|i)=\frac{\sum_{n=1}^N y_{n,i}y_{n,j}}{\sum_{n=1}^N y_{n,i}},\label{eq:caliculate-co-occurence-matrix}
\end{equation}
For example, if the label $j=${\it handwrite} is often attached to a font along with the label $i=${\it script}, $T_{ij}$ becomes large.\par
By using $T$,  we have the completed impression class condition $y^{\rm{CMLE}}_n$ from $y_n$ as follows:
\begin{equation} \label{eq: cases f}
{y^{\rm{CMLE}}_{n, j}}=
    \begin{dcases}
        \frac{\sum_{i=1}^K T_{ij} y_{n, i}}{\sum_{i=1}^K y_{n, i}} & y_{n, j}=0 \\
        1  & y_{n, j}=1
    \end{dcases}
\end{equation}
This equation means that if the $j$-th impression is not attached (i.e., $y_{n,j}=0$), its actual probability is estimated from the attached impressions (i.e., $y_{n,i}=1$) and $T_{ij}$. If the attached impressions have no relevance to $j$-th impression, $T_{ij}\sim 0$ and thus ${y^{\rm{CMLE}}_{n, j}}\sim 0.$ In contrast, if the attached impressions are closely relevant to $j$, ${y^{\rm{CMLE}}_{n, j}}$ becomes a non-zero value. Consequently, CMLE can softly interpolate missing labels according to the co-occurrence relations and remains positive labels as-is. \par
The completed impression class condition ${y^{\rm{CMLE}}_{n, j}}$ can be used as a new ground-truth condition instead of $y_n$. Note again $T_{ij}$ is pre-defined by using the impression labels attached to all $N$ fonts. Therefore,  
we can pre-calculate ${y^{\rm{CMLE}}_{n, j}}$ before training our GAN-model and then use it as a new ground-truth during the training. 
\subsection{Impression label space compressor}
The impression label space compressor (ILSC) compresses the impression label space to a lower-dimensional space for enhancing the robustness to missing labels. Dimensional reduction is a common strategy for data interpolation or data smoothing and has also been used for missing labels~\cite{yu2014large}.  
\par
ILSC is a simple two-layer encoder-decoder. As shown in Fig.~\ref{fig:proposed_model}, its input $y^{\rm Aux}_n$ is the $K$-dimensional impression score vector from the auxiliary classifier, and output $y^{\rm ILSC}_n$ is a $K$-dimensional vector. Specifically, $y^{\rm Aux}_n$ is encoded into a $d (<K)$-dimensional compressed space, and then decoded as $y^{\rm ILSC}_n$. The compressed $d$-dimensional space is spanned by $d$ basis vectors and each of them indicates some group of similar impressions. Consequently, even if the $k$-th impression label is missed in $y^{\rm Aux}_n$, the $k$-th element of $y^{\rm ILSC}_n$ becomes larger than that of the input, by the existence of other impressions similar to $k$.\par

\subsection{Training auxiliary classifier}
The auxiliary classifier outputs a soft prediction of impression probabilities for the generator. As noted above, as the ground-truth of the auxiliary classifier, we do not use $y_n$ but $y^{\rm{CMLE}}_{n, j}$. In addition, instead of using the direct output $y^{\rm Aux}_n$, we use $y^{\rm{ILSC}}_{n, j}$ as the auxiliary classifier output. Combining these two ideas, our model is trained so that $y^{\rm{CMLE}}_{n, j}\sim y^{\rm{ILSC}}_{n, j}$. The vector $y^{\rm{ILSC}}_{n, j}$ is then fed to the generator as a completed impression class condition.
\par
\subsection{Implementation details}
\subsubsection{Perturbing semantic condition}
To have more variations in the generated font images from the semantic condition $s_n$, 
we introduce a Gaussian perturbation of $s_n$. Specifically, we perturb $s_n$ to be $s'_n$ by the reparametarization trick used in variational auto-encoder (VAE), that is, 
$s'_n=\mu(s_n)+\sigma(s_n)\odot \epsilon$, where $\epsilon \sim \mathcal{N}(0,1).$
The functions $\mu(s_n)$ and $\sigma(s_n)$ are realized by a single fully-connected layer and thus this perturbation is fully differentiable, 

\subsubsection{Progressive architecture}
A particular property of font images is that even a small deformation or jaggy in the character contour is very conspicuous. This might be because font images are purely binary images and need an ``artificial'' shape drawn in Bezier curves. Consequently, we need to be more careful of the quality of the resulting images than natural image generation.\par
We, therefore, employ a progressive structure~ \cite{karras2017progressive} for the generator and the discriminator, where the resolution of the generated images increases step by step from $4\times4$ to $64\times64$. The progressive structure realizes high-quality image generation and stable learning.
\subsubsection{Style consistency discriminator}
All 26 character images (from `A' to `Z') of a certain font have consistency in their style. For example, if `A' of a font shows a {\it fat} style, the other 25 characters will also show in the same {\it fat} style. In our case, the 26 characters given by the same semantic condition $s_n$ should have a style consistency.\par
We, therefore, introduce an additional discriminator, called a style consistency discriminator. This discriminator judges whether a pair of its inputs are in the same style or not. There are two types of pairs: consistent pairs and inconsistent pairs.  
A consistent pair is a pair of a fake image and a real image from the $n$th font. Note that the fake image is generated by using $y_n$. An inconsistent pair is a pair of a fake image and  a real image of a randomly-chosen font $n'\neq n$. In practice, a set of $m$ real images is used as the discriminator input to increase stability. \par
\subsubsection{Loss function}
Our model uses three types of losses. The first is an adversarial loss proposed in WGAN-GP\cite{gulrajani2017improved} for the  generator and the discriminator. The second is binary cross-entropy (BCE) loss between the real-valued vectors $y^{\rm CMLE}$ and $y^{\rm ILSC}$. By minimizing this loss, we expect $y^{\rm CMLE}_{n, j}\sim y^{\rm ILSC}_{n, j}$.
The third is Kullback-Leibler (KL) loss as character classification.
\subsubsection{Image generation}
When we generate a font image by a (one-hot or multiple-hot) impression vector $y$, the vector $y$ is first transformed by CMLE, and the resulting vector is directly used  as a weight to have the semantic vector $s$. By feeding $s$, a one-hot character class condition, and a random noise  input to the generator, we have an image with the impression $y$.  
\section{Experimental Results}
\subsection{Experimental setup}
Our experiments were conducted under the following conditions.
The proposed model is trained for 90,000 iterations, and the resolution in progressive increases every 15,000 iterations. Adam~\cite{kingma2014adam} is used as the optimizer with a learning rate of 0.0005. The batch size is 512, and the update ratio of Generator and Discriminator is 1:1. The dimension of the compressed space in ILSC is $d=300$, and the number of reference images in the style consistency discriminator is $m=4$. As noted in Section~\ref{sec:dataset}, $N=17,202$ fonts and $K=1,430$ impression labels were used.

\subsection{Comparative methods}
We conduct quantitative and qualitative evaluations to compare the quality of the generated images by the proposed method with those by the previous methods. The competitors include C-GAN~\cite{mirza2014conditional}, AC-GAN~\cite{odena2017conditional}, CP-GAN~\cite{kaneko2019CP-GAN}. Since their original formulations do not have a module for deriving semantic conditions, we have slightly extended them for fair comparisons. Hereafter, we call their extended models C-GAN+, AC-GAN+, and CP-GAN+, respectively.  We also compare our method with  
Imp2Font~\cite{matsuda2021impressions2font}, which is a GAN-based method specialized for generating font images with specific impressions. We also conduct an ablation study, where the proposed components, CMLE and ILSC, are removed one by one from the entire model.
\subsection{Learned correlation among impression labels}
\begin{figure}[t]
    \centering
    \includegraphics[width=1.0\linewidth]{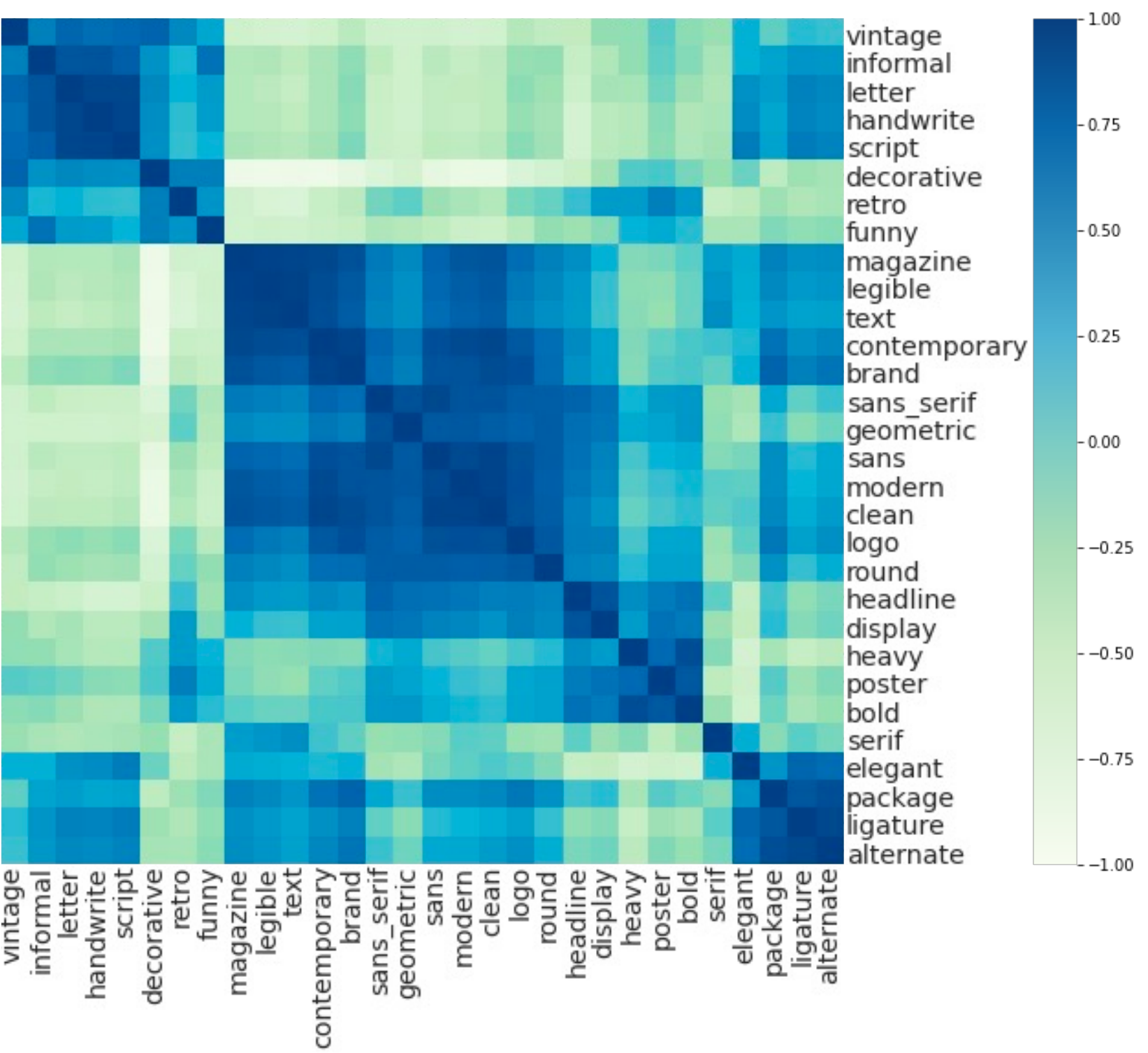}\\[-3mm]
    \caption{Learned correlations between impressions.}
    \label{fig:classifier_correlation_matrix}
\end{figure}

If the two components, CMLE and ILSC, work properly to deal with missing labels, similar impressions should co-occur with each other in $y^{\rm ILSC}$ even though the ground-truth condition $y$ contains missing labels. Fig.~\ref{fig:classifier_correlation_matrix} shows the visualization of the learned correlation among 30 elements in $y^{\rm ILSC}$, where each element corresponds to one of the top-30 frequent impression labels. For the visualization, this correlation matrix is pre-processed by a biclustering-based matrix reordering technique so that the matrix becomes block-diagonal. From this visualization, we can see that similar impressions are properly co-occurring. For example, impression labels such as {\it letter}, {\it handwrite}, and {\it script} have high positive correlations. This is because these impression labels are common to fonts with italic and writing styles. Consequently, if the impression {\it letter} is missed in $y$, it will be boosted in $y^{\rm ILSC}$ with the help of {\it handwrite} and {\it script} in $y$.
\begin{table}[t]
 \caption{Quantitative evaluation results of generated images.}
 \label{table:quantitative_evaluation}\vspace{-2mm}
 \centering
{\addtolength{\tabcolsep}{-1mm}
  \begin{tabular}{lrrrr}
  \hline
   & $\downarrow$FID  & $\downarrow$Intra-FID  &  {\scriptsize $\uparrow$mAP-train\red{(\%)}}& {\scriptsize$\uparrow$ mAP-test\red{(\%)}}\\ \hline 
    C-GAN+\cite{mirza2014conditional} & 29.618 & 52.199 &  1.321& 1.390\\ 
    AC-GAN+\cite{odena2017conditional} & 29.152 &  68.355  & 1.189 & 1.321\\ 
    CP-GAN+\cite{kaneko2019CP-GAN} & 30.412 & 152.398 & 1.366 & 1.402\\ Imp2Font\cite{matsuda2021impressions2font} & 24.543 & 146.691 & 1.284& 1.454 \\ \hline 
    w/o ILSC & 26.254 &  63.062 & 1.426 & 1.322 \\ 
    w/o CMLE & 26.942 &  \bf{50.194} & 1.370  & 1.459   \\ \hline 
    \bf{Full model} & \bf{21.895} & 56.733 & \bf{1.638}  & \bf{1.806}\\ \hline 
  \end{tabular}
  \vspace{-0.5cm}
}
\end{table}
\begin{figure}[t]
    \centering
    \includegraphics[width=1.0\linewidth]{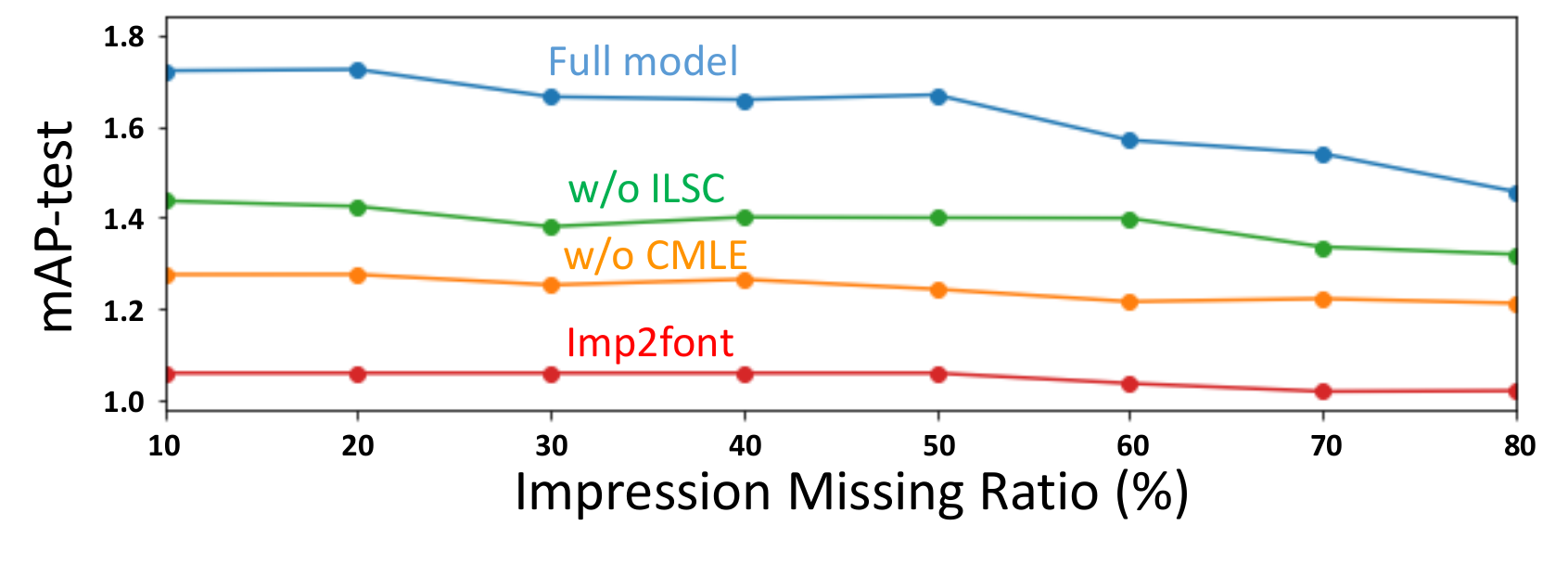}\\[-5mm]
    \caption{mAP-test under different ratios of  missing labels. }
    \label{fig:mAP_test_impression_missing_ratio}
    \vspace{-2mm}
\end{figure}

\subsection{Quantitative evaluation}
In the quantitative evaluation, four metrics are used: FID~\cite{heusel2017gans}, Intra-FID~\cite{miyato2018cgans}, mAP-train, and mAP-test. 
\begin{itemize}
    \item FID measures the diversity and quality of the generated images by using the pre-trained Inception network. We calculate FID using $26\times 5,000$ samples,  where $5,000$ samples are generated using the randomly-selected impression class conditions.
    \item 
    Intra-FID is the average of FID calculated for each impression class. Since Intra-FID needs enough samples for each class, we only use the frequent impression classes attached to more than 200 fonts.
    \item mAP-train and mAP-test are metrics, extended from  GAN-train~\cite{shmelkov2018good} and GAN-test~\cite{shmelkov2018good} for dealing with multi-label conditions. mAP-train is the mean average precision (mAP) of the classification results of real images using the classifier trained on generated images. mAP-test is the mAP of the classification results of the generated images using a classifier trained on real images.  mAP-train and mAP-test mainly evaluate the diversity and  the quality of generated images, respectively. In our experiments, ResNet50 is used as the classification model. To calculate the mAP, we use the same method described in Chen et al.~\cite{Chen2019large}.
\end{itemize}
\par

Table~\ref{table:quantitative_evaluation} shows the results of the quantitative evaluation. The results show that the full model significantly outperforms all other models in most metrics. Specifically, through the comparison with the ablation models, we show that the introduction of CMLE and ILSC had a significant impact on improving the diversity and quality of the conditional generating distribution. 
In particular, the full model significantly improves the mAP-test.
This metric measures whether the style features of the font are properly captured or not. The missing labels make learning the relationship between appropriate impressions and style difficult. Therefore, this mAP-test improvement shows that the effect of missing labels can be reduced. The full model is only inferior to  ``w/o CMLE'' by intra-FID. This is because 
CMLE has a smoothing effect of impressions and the effect of the original impressions becomes slightly weaker.\par

Fig.~\ref{fig:mAP_test_impression_missing_ratio} shows the mAP-test values under different impression missing ratios. In this experiment, a certain ratio of the original impression labels are intentionally removed when we make fake images for mAP-test. The proposed method achieves a higher mAP-test than the comparative methods at any ratios. This result shows that the proposed method is the most robust against missing labels.

\subsection{Qualitative evaluation}
\begin{figure*}[t]
    \centering
    \includegraphics[width=1.0\linewidth]{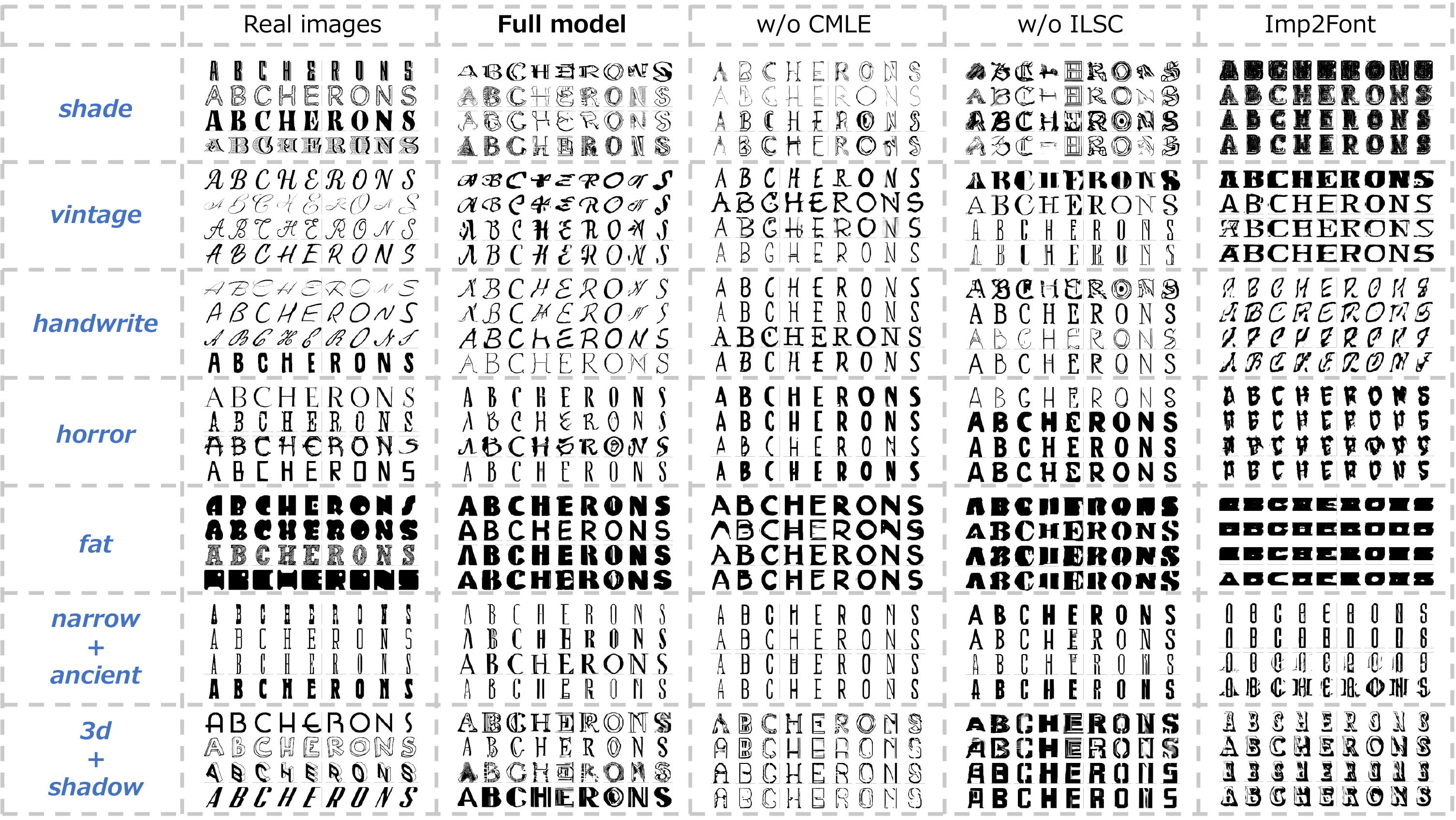}\\[-3mm]
    \caption{Generated images by specifying impression labels. Real images are randomly selected from fonts that have the specified impression labels.}
    \label{fig:generated_images_from_specific_impressions}
    \vspace{-0.5cm}
\end{figure*}

\begin{figure}[t]
    \centering
    \includegraphics[width=1.0\linewidth]{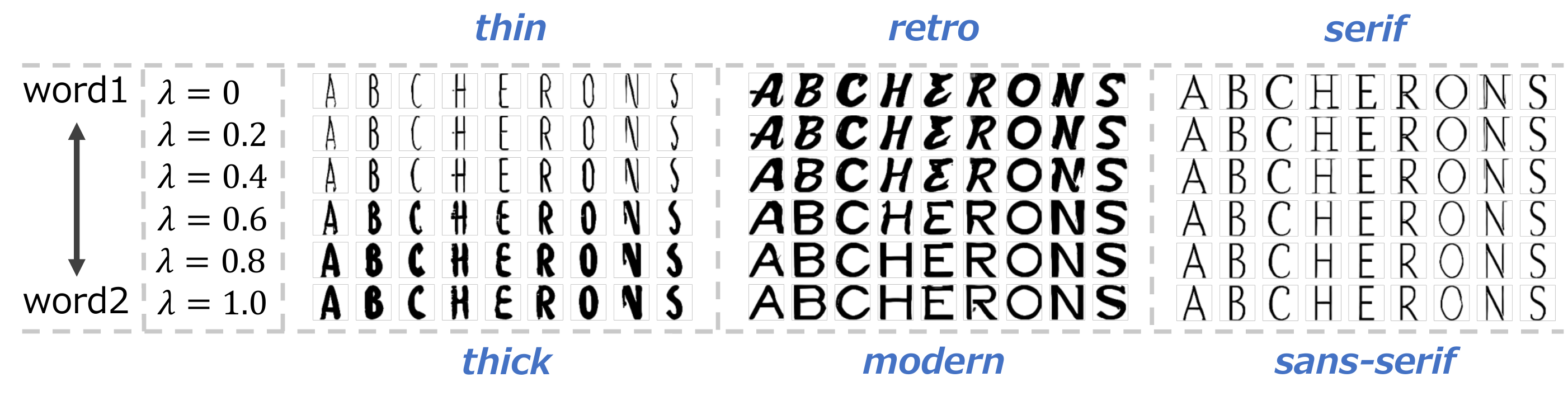}\\[-3mm]
    \caption{Font images generated while interpolating two impression labels.}
    \label{fig:interpolation_impression}
\bigskip
    \includegraphics[width=1.0\linewidth]{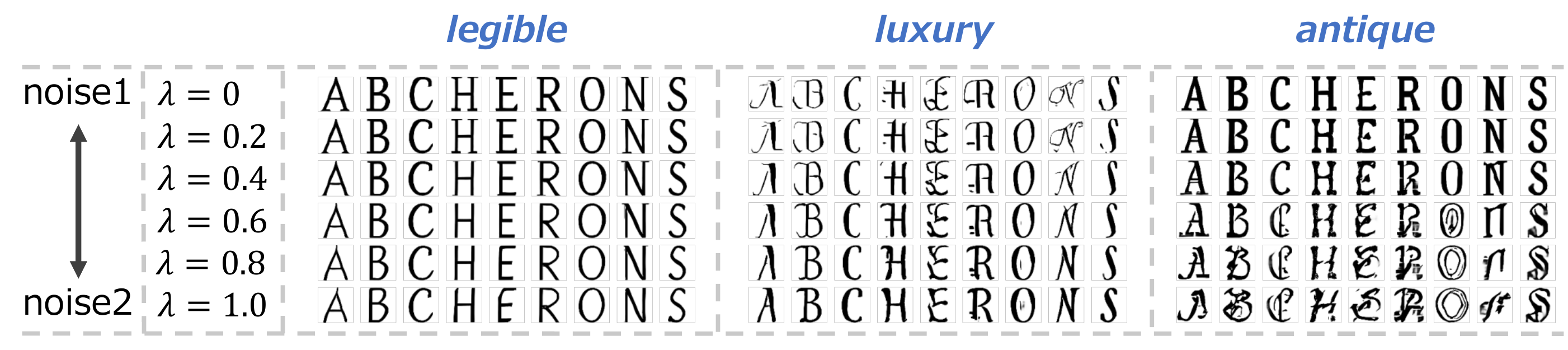}\\[-3mm]
    \caption{Font images generated while interpolating two different noise inputs.}
    \label{fig:interpolation_noise}\vspace{-3mm}
\end{figure}

Fig.~\ref{fig:generated_images_from_specific_impressions} shows the fonts generated from a single impression label (the top 5 rows) or a pair of impression labels (the bottom two rows). We used the fixed string ``ABCHERONS'' to observe the generated font images~\footnote{The string ``HERONS'' is often used to observe font styles because it contains most of all stroke shapes in Latin capital alphabets.}. 
As comparative methods, we used two ablation models (``w/o CMLE'' and ``w/o ILSC'') and Imp2Font, which achieved the best performance in the quantitative evaluation among the comparative methods.\par
Fig.~\ref{fig:generated_images_from_specific_impressions} shows that the generated images show the specified impressions properly. For example, the full model can generate images that inherit large curls of real images for {\it vintage} impression, whereas other methods cannot. Moreover, the full model can generate a larger variety of font styles from the same impression label, compared to the other methods. Having large variations indicates that the proposed method is more appropriate for a font design tool. Note that the existence of the large variations is supported by the above quantitative evaluation result that the full model has the highest mAP-train. 
\par

Our method is also suitable for generation from multiple impression labels. Especially for {\it narrow + ancient}, although almost all the models realize narrow characters, only the full model realizes the ``serif,'' which is often found in the fonts with {\it ancient} impression.  \par

Fig.~\ref{fig:generated_images_from_specific_impressions} also shows the comparison with the generated fonts by Imp2Font~\cite{matsuda2021impressions2font}. We can see that there is less diversity as a font generated by Imp2font. Also, the readability of the font generated by Imp2font is very low (see, e.g. {\it fat}). Meanwhile, the proposed method captures the styles represented by impression labels, and at the same time, it shows high diversity and readability.\par

The proposed model enables us to control the strength of the impression labels. For example, we can use the impression class condition by an interpolation between two different impression labels. For example, the interpolation between  {\it thin} and {\it thick} is represented by $y_{{\it thin}\leftrightarrow{\it thick}}= (1-\lambda)y_{\it thin} + \lambda{y_{\it thick}}$. 
The generated images by several interpolated conditions are shown in Fig.~\ref{fig:interpolation_impression}. Note that we fixed the noise $z$ in this experiment. It is shown that our method can generate appropriate intermediate font shapes. 
\par
It is also possible to interpolate two different noise inputs while fixing the impression class condition.  The interpolated result is given as $z= (1-\lambda)z_{1} + \lambda{z_{2}}$. Fig.~\ref{fig:interpolation_noise} shows images generated by a certain interpolation coefficient $\lambda$. The results show that the choice of noise can yield a variety of fonts with specific common styles. \par

\section{Conclusion and Future Work}
This paper develops a new conditional GAN robust to missing labels and applies it to font generation with specific impression labels. We introduce two components to deal with missing labels: co-occurrence-based missing label estimator (CMLE) and impression label space compressor (ILSC). Quantitative evaluations prove the proposed method can generate higher quality and more diverse font images than the existing methods, under multiple evaluation metrics. Qualitative evaluations also support that the proposed method can generate robust images for missing labels.  \par
Future work will focus on several applications. First, we can use the current method for a font impression estimator. The auxiliary classifier in the current model is already trained for this application. Second, we can analyze ILSP to understand the redundancy of the impression classes. Last but not least, we will utilize the proposed model for understanding the relationship between font shape and its impressions.  

\section*{Acknowledgment}
This work was partially supported by JSPS KAKENHI Grant Number JP17H06100.




\newpage
\bibliographystyle{IEEEtran}
%


\bibliography{icpr}


\par
\vphantom{hoge}\par

\section*{Appendix}

\setcounter{figure}{0}
\setcounter{table}{0}
\setcounter{section}{0}
\renewcommand{\thesection}{\!A\arabic{section}}
\renewcommand{\thefigure}{\!A\arabic{figure}}
\renewcommand{\thetable}{\!A\arabic{table}}
\renewcommand{\topfraction}{.9}
\renewcommand{\bottomfraction}{.9}
\renewcommand{\textfraction}{.01}
\renewcommand{\floatpagefraction}{.9}

\section{Details of impression labels and missing labels}
The MyFonts dataset is composed of fonts collected from {\tt Myfonts.com}. Each font is originally attributed by its font family, designer, and foundry. In addition to those attributes, each font is tagged with various words, such as {\it serif}, {\it elegant}, and {\it pretty}. We call these tags impression labels. Roughly speaking, there are three types of impression labels. The first type refers to a typical font style, such as  {\it sans-serif} and {\it script}. The second refers to a more shape-related property, such as {\it bold} and {\it oblique}.
The third refers to a more abstract impression, such as {\it elegant} and {\it scary}. In this paper, we simply called them  impression labels. \par
As noted in Section~\ref{sec:intro}, the impression labels are often attached by non-experts. Moreover, the impression labels are open-vocabulary; there is no pre-defined list of impression labels. These two conditions incur  {\it noisy} impression labels, which are very personal and not commonly-acceptable labels.\par
Moreover, under the above conditions, it is difficult to expect that all appropriate impression labels are attached to each font. Accordingly, we often find {\it missing} impression labels. For example, for two very similar fonts, an impression label may be attached to only one of the fonts, but not the other. In this case, the latter font has (at least one) missing label.\par
%

\section{Detailed illustrations of ILSC and CMLE}
\begin{figure}[!b]
    \centering
    \includegraphics[width=0.75\linewidth]{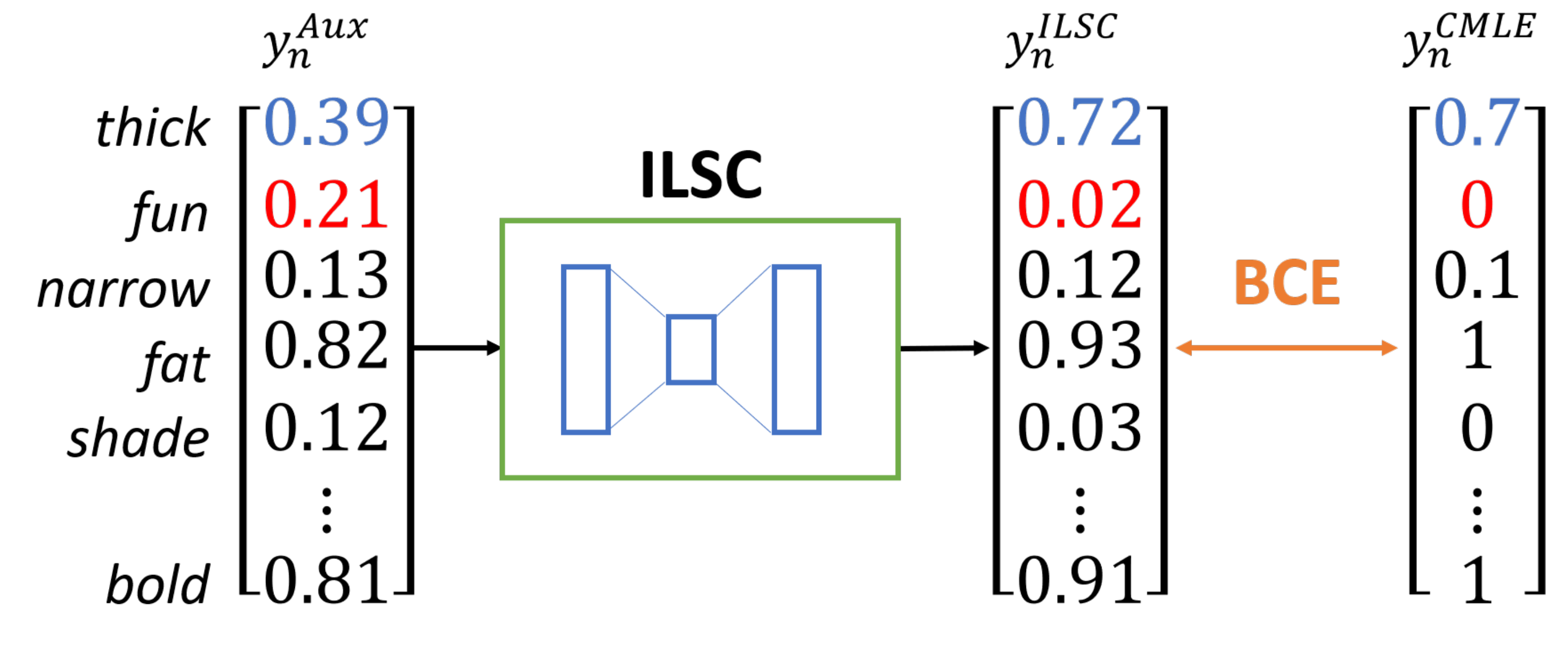}
    \caption{Function of impression label space compressor (ILSC).\label{fig:ILSC}}
\bigskip
    \includegraphics[width=\linewidth]{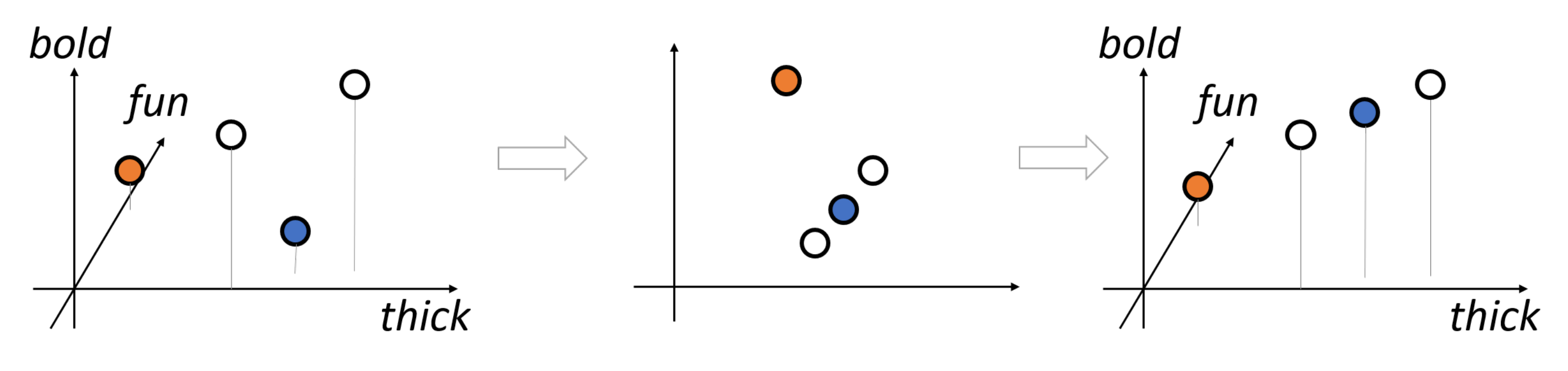}
    \caption{Mechanism of ILSC.\label{fig:ILSC2}}
   
\end{figure}

\begin{figure}[!t]
    \centering
 \includegraphics[width=0.75\linewidth]{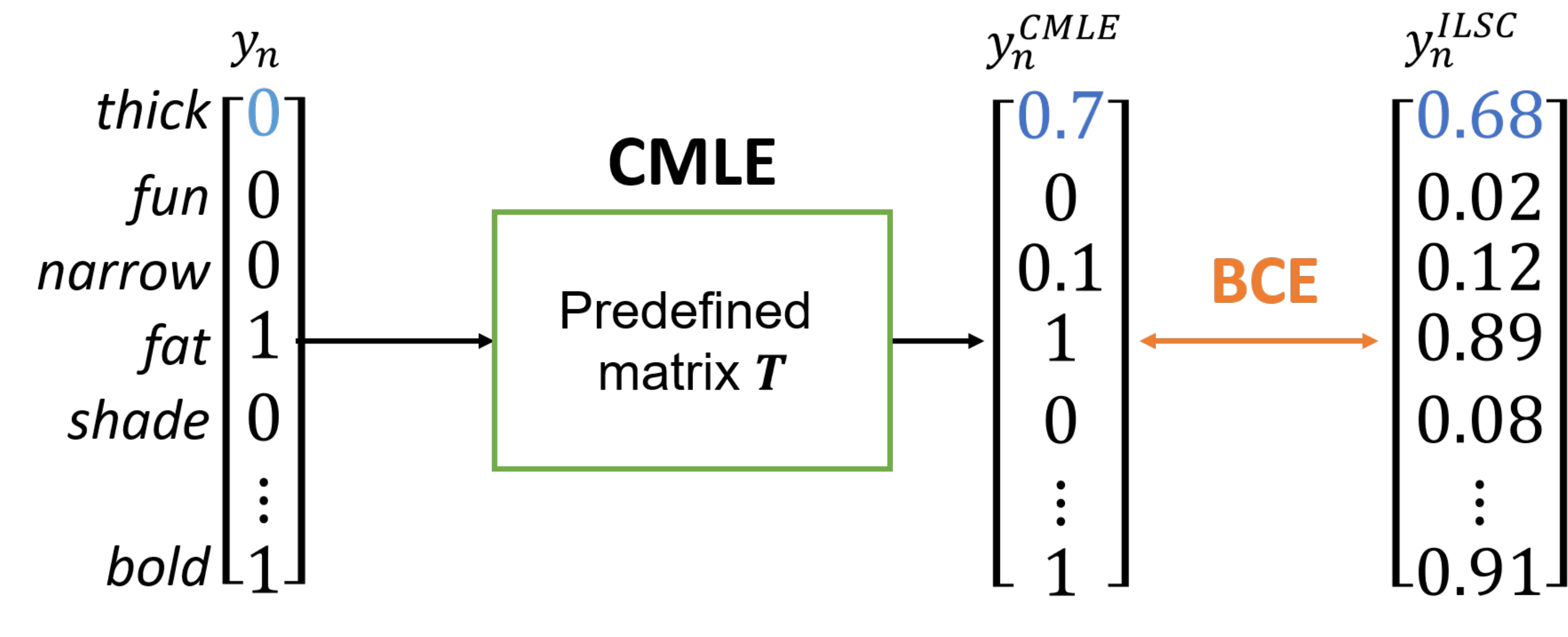}
    \caption{Function of co-occurrence-based missing label estimator (CMLE).\label{fig:CMLE}}
\bigskip
    \includegraphics[width=\linewidth]{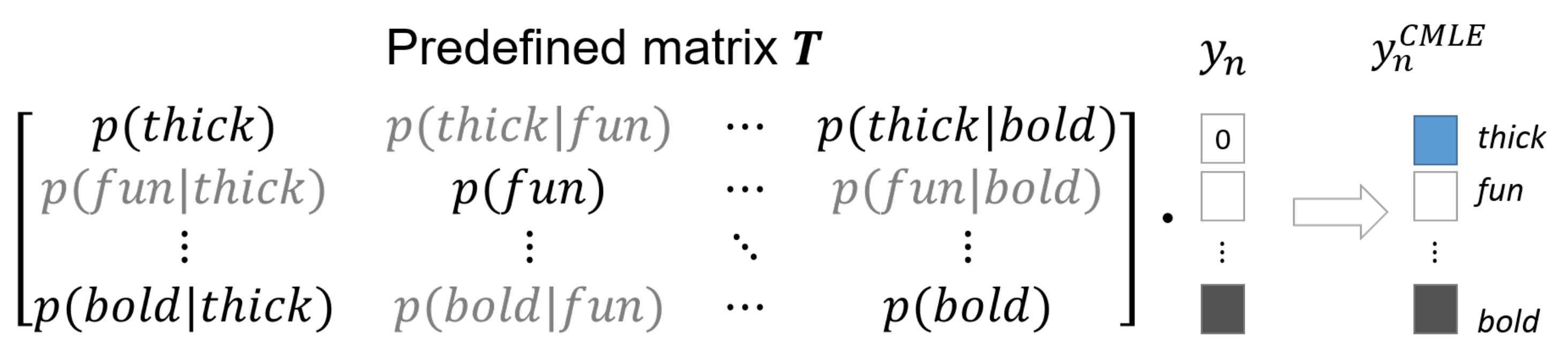}
    \caption{Mechanism of CMLE.\label{fig:CMLE2}}
\end{figure}

{Figs.~\ref{fig:ILSC}} and {\ref{fig:ILSC2}} illustrate the details of ILSC. As shown in these figures, ILSC tries to compensate for the missing labels (see a blue dot in 
{Fig.~\ref{fig:ILSC2}}, where {\it bold} is missed) by embedding to a low-dimensional subspace.
Similarly, {Figs.~\ref{fig:CMLE}} and {\ref{fig:CMLE2}} illustrate the details of CMLE. {Fig.~\ref{fig:CMLE2}} shows that CMLE  compensates for the missing labels by using the predefined co-occurrence matrix $T$.\par
\par

\section{Details of mAP evaluation}
One might think that our mAP is too low, compared to, for example, PASCAL VOC object detection experiments, where we can often see mAP values around 10$\sim$30\%. However, there are clear reasons why our mAP becomes around 1\%.\par
Before explaining the reasons, let us review the calculation of mAP. mAP is the mean of average precisions (APs) for individual impression labels. Assume that there are $H$ fonts $\{X_1, X_2,..., X_H\}$ with the impression label $i$ in the test set. Also, assume that all 17,202 fonts are sorted by their likelihood of the impression label $i$. (Note that the likelihood is given by a CNN-based  impression classifier that classifies fonts into positive and negative classes about their impression $i$.)  Then, we have the rank of $H$ fonts in the sorted result as $\{r_1, r_2,..., r_H\}$. Using this rank, the AP of $i$ ($\mathrm{AP}_i$) is given as follows:
\begin{equation}
    \mathrm{AP}_i = \frac{1}{H}{\sum_{h=1}^H}\frac{h}{r_h}.
\end{equation}
Then, mAP is given as the mean value of the APs for all impression labels, that is, $\mathrm{mAP} = \sum_{i=1}^K\mathrm{AP}_i$.\par
\begin{figure}[t]
    \centering
    \includegraphics[width=1.0\linewidth]{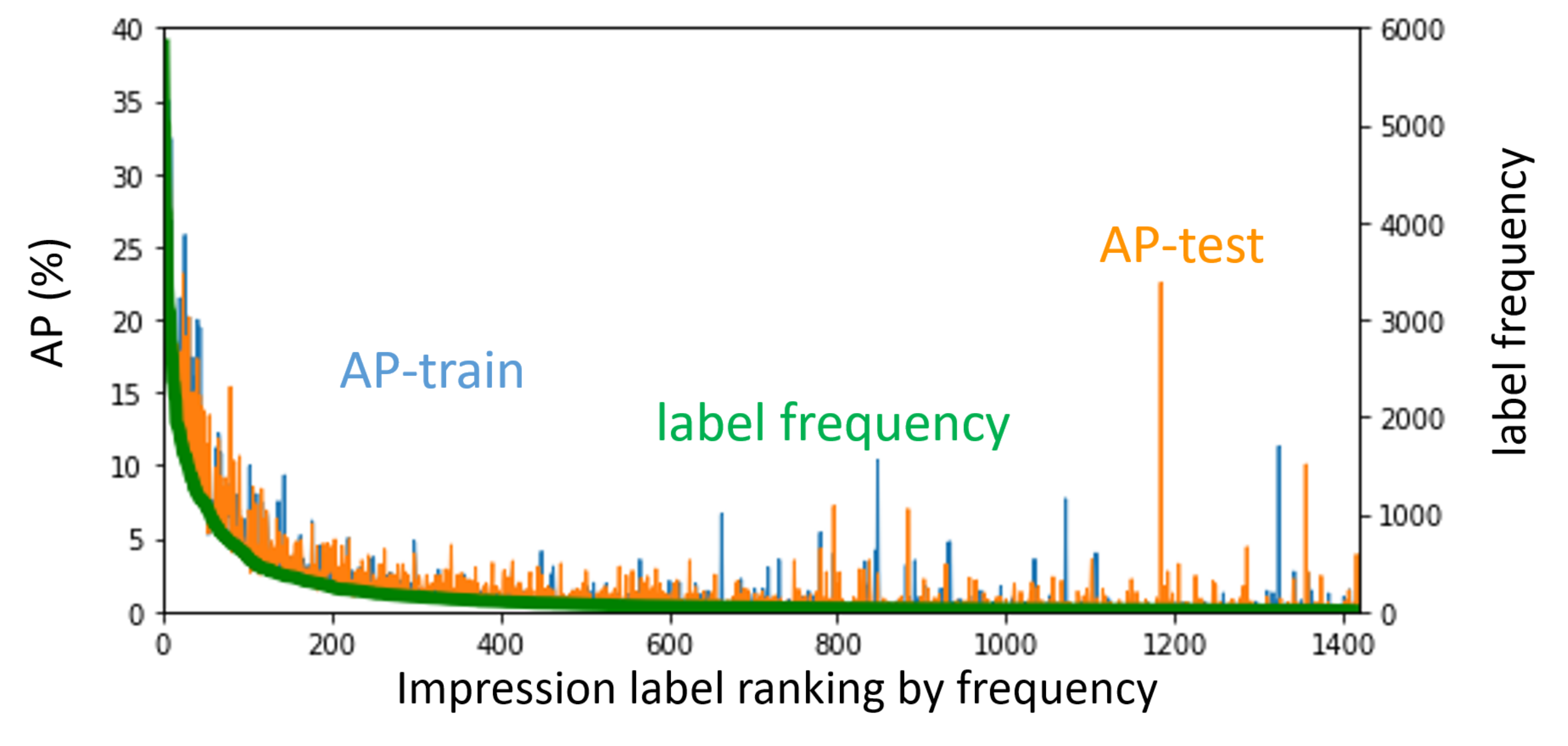}
    \caption{Frequency, AP-train, and AP-test for each impression label.} \label{fig:AP}
\end{figure}
The reasons why our mAP is low are as follows. The first reason is many missing labels. As noted above, even appropriate impression labels are often missed. For example, a funny font $X$ will not have the {\it funny} label. In this case, even if $X$ is highly ranked, it is ignored -- in other words, it does not contribute to boosting the AP value for {\it funny}. The second and more serious reason is that there are many minor impression labels. As noted above, impression labels are open-vocabulary and many people use arbitrary words as impression labels. The AP values of those minor impression labels tend to be low. For an extreme example, if only a single font has the $i$th impression and it is ranked at \#100 among 17,202 fonts, $\mathrm{AP}_i = 1\%.$\par
{Fig.~\ref{fig:AP}} shows AP-train/test and frequency of each of 1,430 impression labels. The labels are sorted by their frequency. This result clearly shows that there are so many minor impression labels and their AP values tend to be very low. Consequently, the ``mean'' AP over all impressions becomes around 1\%. Note that if we take the mean of AP values of top-$M$ frequent impressions, mAP-train (mAP-test) becomes 29.0 (29.5), 16.7 (16.7), and 11.7 (12.0) for $M=10, 50$, and $100$, respectively. This observation proves that our model generates very appropriate fonts for frequent impressions.\par
It is possible to remove minor impressions from our experiments --- however, we did not. This is because we can still compare the methods in Table~\ref{table:quantitative_evaluation} if these methods are evaluated under the same condition (including minor impressions). In addition, we could avoid a hyper-parameter, i.e., the threshold to cut out minor impressions. Since there is no objective way to set the threshold, the results with the threshold become somewhat subjective.  

\section{Details of binary cross entropy}
We used the standard binary cross-entropy (BCE) for two real-valued vectors, $y_{n,i}^{\rm CMLE}$ and $y_{n,i}^{\rm ILSC}$. (Usually, one of them is a binary vector.)  More formally, our BCE loss is written as: 
\begin{eqnarray*}
     L_{\mathrm{BCE}}&=& -{\sum_{n=1}^N}{\sum_{i=1}^K} y_{n,i}^{\rm CMLE}\log{y_{n,i}^{\rm ILSC}}\\
     && +(1-y_{n.i}^{\rm CMLE})\log(1-{y_{n,i}^{\rm ILSC}}).
\end{eqnarray*}
By minimizing them, we can expect  $y_{n,i}^{\rm CMLE}\sim y_{n,i}^{\rm ILSC}$. We revised the main text to clarify that our BCE is calculated between the two real-valued vectors, $y_{n,i}^{\rm CMLE}$ and $y_{n,i}^{\rm ILSC}$.

\section{Readability of generated fonts}
Although we have conducted an mAP-based quantitative evaluation, the evaluation of readability is, actually, not straightforward. In this work, we deal with various fonts with various impressions. Among those fonts, even ``real'' ones are often hard to read. For example, the bottom example of {\it fat} fonts in Fig.~\ref{fig:generated_images_from_specific_impressions} shows hard readability. Moreover, similar to impressions, readability is subjective and will show heavy reader-dependency.
\par
In near future, we will try to develop a ``(readable) character versus (unreadable) non-character'' classifier by using some 1-class classifiers. We also will try to establish  a subjective evaluation protocol for the readability of those various fonts. These trials are, actually, very interesting to understand what the letters are.

\end{document}